\documentclass{article}

\usepackage[preprint]{neurips_2025}

\usepackage[utf8]{inputenc} 
\usepackage[T1]{fontenc}    
\usepackage{hyperref}       
\usepackage{url}            
\usepackage{booktabs}       
\usepackage{amsfonts}       
\usepackage{nicefrac}       
\usepackage{microtype}      
\usepackage{xcolor}         

\usepackage{algorithm}
\usepackage{algorithmic}
\usepackage{comment}
\usepackage{natbib}
\usepackage{graphicx}
\usepackage{adjustbox}
\usepackage{multirow}
\usepackage{tikz}
\usepackage{makecell}
\usepackage{todonotes}
\usepackage[font=small,labelfont=bf]{caption}

\title{\textsc{Gala:} Global LLM Agents for \\ Text-to-Model Translation}

\author{
  Junyang Cai$^{1}$, 
  Serdar Kad{\i}o\u{g}lu$^{2,3}$, 
  Bistra Dilkina$^{1}$ \\
  $^{1}$Department of Computer Science, University of Southern California \\
  $^{2}$AI Center of Excellence, Fidelity Investments \\
  $^{3}$Department of Computer Science, Brown University \\
  \texttt{caijunya@usc.edu}, 
  \texttt{serdark@cs.brown.edu}, 
  \texttt{dilkina@usc.edu}
}

\begin{document}

\maketitle

\begin{abstract}
Natural language descriptions of optimization or satisfaction problems are challenging to translate into correct \textsc{MiniZinc} models, as this process demands both logical reasoning and constraint programming expertise. We introduce \textsc{Gala}, a framework that addresses this challenge with a global agentic approach: multiple specialized large language model (LLM) agents decompose the modeling task by global constraint type. Each agent is dedicated to detecting and generating code for a specific class of global constraint, while a final assembler agent integrates these constraint snippets into a complete \textsc{MiniZinc} model. By dividing the problem into smaller, well-defined sub-tasks, each LLM handles a simpler reasoning challenge, potentially reducing overall complexity. We conduct initial experiments with several LLMs and show better performance against baselines such as one-shot prompting and chain-of-thought prompting. Finally, we outline a comprehensive roadmap for future work, highlighting potential enhancements and directions for improvement.
\end{abstract}

\section{Introduction}
Building correct \textsc{MiniZinc} models from natural language descriptions is a complex challenge. Recently, \cite{singirikonda2025text2zinc} has introduced several modeling co-pilot frameworks as well as a supporting dataset, \textsc{Text2Zinc}, and corresponding leaderboard for benchmarking this task. Existing co-pilot evaluations on \textsc{Text2Zinc} (using direct prompting, chain-of-thought, and compositional strategies, and external tools such as \textsc{Ner4Opt}~\citep{ner4opt2024}, Knowledge Graphs) found that even powerful LLMs are “not yet a push-button technology” for generating combinatorial models from text. In other words, general-purpose prompting often fails to capture all variables and constraints correctly, especially for harder optimization problems. This motivates research into more structured and guided methods.

One promising direction is to break the problem into manageable pieces. Multi-step or multi-agent frameworks have begun to emerge for NL4Opt~\citep{ramamonjison2023nl4opt} (natural language for optimization) tasks. For example, Chain-of-Experts by ~\cite{xiao2023chain} assigns multiple LLM “experts” with specific roles (e.g. interpreting text, formulating model components, coding, verifying) coordinated by a central conductor. This cooperative agent approach significantly outperformed prior single-LLM methods on complex operations research problems. Similarly, the OptiMUS system by \cite{ahmaditeshnizi2023optimus} uses an LLM-based agent to iteratively identify parameters, write constraints, and debug a linear program model, achieving a higher problem-solving rate than basic one-shot prompting. These results suggest that decomposing the modeling task and giving LLMs more structured guidance can substantially improve performance.

The main drawback of existing multi-agent approaches is that each agent still inherits the full complexity of the problem rather than focusing on a narrower, more tractable sub-task. To address this, we propose a new agentic framework for translating text to \textsc{MiniZinc}~\citep{netherCoTe2007minizinc} that is centered around global constraints. In Constraint Programming (CP), global constraints such as  \texttt{all\_different} and \texttt{cumulative} are high-level primitives that capture common patterns among variables. They offer expressive and efficient building blocks in models, and many optimization problems can be described as a combination of such global constraints. 

Our approach, \textbf{\textsc{Gala}}: \textbf{G}lob\textbf{A}l \textbf{L}LM \textbf{A}gents leverages this by dedicating a specialized LLM agent to each type of global constraint, turning model generation into a collaboration of focused experts rather than one monolithic generation. More broadly, our approach can be viewed as aligning and combining the key strength of Constraint Programming (i.e., global constraints) with the Agentic Frameworks. The following sections outline background and related work, our framework, preliminary findings, and future roadmap. 

\section{Background and Related Work}
\vspace{-0.1cm}
\subsection{\textsc{Holy Grail 2.0}, \textsc{NL4Opt} \textsc{Ner4Opt}, \& \textsc{Optimus}}
\vspace{-0.1cm}
LLMs are increasingly applied to optimization and constraint programming. Holy Grail 2.0~\citep{tsouros2023holy} outlined a blueprint for conversational modeling assistants. Early work \cite{ramamonjison2023nl4opt} tackled linear programming via entity recognition and logical-form translation, showing promising ChatGPT results on NL4OPT.\textsc{Ner4Opt} showed that accuracy of LLM-generated \textsc{MiniZinc} improves with in-line entity annotations~\citep{dakle2023ner4opt} as also built into several co-pilot pipelines~\citep{ner4opt2024}. Agentic methods followed: a multi-agent Chain-of-Experts~\citep{xiao2023chain} and Optimus~\citep{ahmaditeshnizi2023optimus}, a modular system for complex descriptions. Text-to-model translation was also explored via simple decomposition prompts with GPT~\citep{tsouros2023holy}. Extending this, RAG-based in-context learning built CPMpy models~\citep{michailidis2025cp}. Concurrently to our work, \cite{szeider2025cp} introduce a Reason-and-Act framework solving all 101 CP-Bench tasks~\citep{michailidis2025cp}.

\vspace{-0.1cm}
\subsection{\textsc{MiniZinc} \& \textsc{Text2Zinc}}
\vspace{-0.1cm}
\textsc{MiniZinc}~\citep{netherCoTe2007minizinc} is a high-level, solver-agnostic constraint modeling language for discrete and continuous satisfaction and optimization problems. It provides a rich library of global constraints that capture common CP patterns, allowing users to express problems declaratively rather than through low-level decompositions. \textsc{MiniZinc}’s practical utility is enhanced by its clean separation between models and instances. Given these features and the availability of existing \textsc{Text2Zinc}\citep{singirikonda2025text2zinc} datasets, we benchmark different approaches of NL4OPT on \textsc{MiniZinc} execution accuracy and solution accuracy.

\vspace{-0.1cm}
\subsection{CP \& Global Constraints}
\vspace{-0.1cm}
In CP, global constraints concisely represent recurring patterns like all-differentness, resource limits, ordering, counting, etc., which are found across many scheduling, assignment, and configuration problems. For example, \texttt{all\_different} enforces that a set of decision variables all have distinct values – a common requirement in scheduling and allocation problems. Other examples include \texttt{cumulative} (which enforces scheduling resource capacity over time) and \texttt{global\_cardinality} (which limits how many variables take each value). Our approach \textsc{Gala} is designed to take advantage of CP global constraints in an agentic LLM architecture as detailed next.

\section{\textsc{Gala}: Global LLM Agents}
\textbf{Specialized LLM Agents per Constraint:} For each type of global constraint, we instantiate a separate LLM agent with a specialized prompt. This prompt primes the agent to act as an expert in detecting and formulating that constraint in \textsc{MiniZinc}. The agent receives the full problem description (and any input data) but its instructions are local: it should ignore the broader context and only answer the question, “Does this problem involve an <X> constraint? If yes, produce the \textsc{MiniZinc} snippet for it; if not, output FALSE.” Each agent is effectively performing a binary classification (constraint present or not) followed by code generation for that constraint if needed. 

Critically, the agent is instructed not to produce any other modeling elements beyond its constraint. For instance, the \texttt{all\_different} agent is prompted: “You are a \textsc{MiniZinc} modeling assistant specialized in detecting and modeling \texttt{all\_different} constraints. Given a problem description, decide whether it requires one or more \texttt{all\_different} constraints. If it does, generate only \textsc{MiniZinc} code specifying the \texttt{all\_different} constraint and its variables. If it does not, return FALSE with a brief reason.”. Similar templates are crafted for each global constraint type, incorporating definitions and common clue words (e.g. phrases like "each ... different" hint at \texttt{all\_different}, or "no overlap" hints at \texttt{cumulative}). By isolating each agent’s focus, \textbf{our main novelty is to  simplify the reasoning task leveraging CP global constraints}. \textit{Unlike the previous agentic approaches, our agents do not need to understand the entire problem structure, only whether a specific pattern appears and how to encode it.}

\textbf{Assembler LLM for Model Integration:} Once the constraint-specific agents have each returned either a code snippet or FALSE, an assembler LLM takes over. The assembler’s input includes the original problem description and all the constraint snippets (we call them ``hints'') provided by the agents that found a constraint. The assembler’s role is to compile a complete, coherent \textsc{MiniZinc} model from these pieces. We prompt the assembler agent as if it were “the world’s best \textsc{MiniZinc} programmer” tasked with integrating hints and filling in the gaps. Concretely, it must: (1) declare all decision variables (and their domains) that are needed, possibly renaming or merging variables from different snippets for consistency; (2) analysis and decide if to include provided global constraints or not; (3) add any remaining constraints from the text that were not covered by the hints; (4) determine the objective (if an optimization problem) or a satisfy goal, based on the description; and (5) append a proper solve item and output format. The assembler may ignore an irrelevant hint if an agent was mistaken, but in general, it tries to use all valid snippets. By design, this agent has the most complex task, since it sees the full problem and must ensure completeness. However, because much of the heavy lifting (expressing complex constraints) is done by the specialized snippets, the assembler can focus on glueing components together and writing boilerplate code.

\begin{figure}[t]
\centering
\begin{minipage}[t]{0.48\linewidth}
  \centering
  \includegraphics[width=\linewidth]{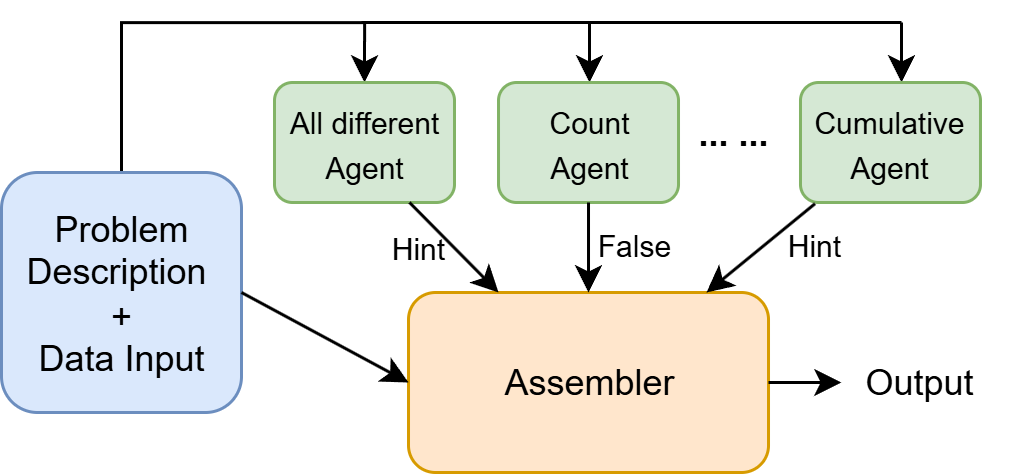}
  \captionof{figure}{\textsc{Gala}: Global LLM Agents architecture for global-constraint detection and assembly.}
  \label{fig:left}
\end{minipage}\hfill
\begin{minipage}[t]{0.48\linewidth}
  \centering
  \includegraphics[width=\linewidth]{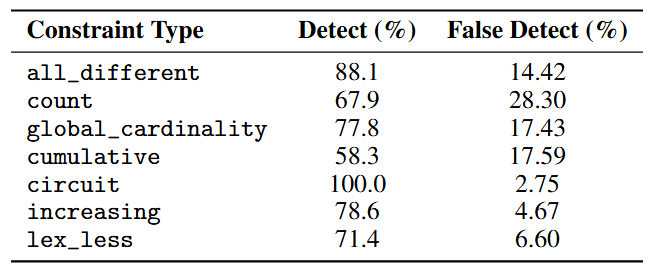}
  \captionof{table}{Performance of \textsc{Gala} on detection and false-detection rates by global constraint type.}
  \label{tab:constraint_metrics}
\end{minipage}
\end{figure}

Figure~\ref{fig:left} presents the overall architecture which is essentially a team of specialists: each constraint agent \textit{independently} proposes part of the model, and the assembler (acting as the principal architect) reviews and integrates these contributions into the final solution. This approach aims to reduce the cognitive load on any single LLM. Eech agent works on a local model of the structured sub-problem, thanks to CP global constraints, instead of the entire text-to-model translation.


\vspace{-0.1cm}
\section{Initial Results}
\vspace{-0.1cm}
We conduct an initial evaluation of \textsc{Gala}, focusing on two aspects: (1) the ability of global agents to correctly detect global constraints, and (2) the end-to-end performance of the agentic pipeline compared to baseline prompting strategies and chain-of-thought (CoT)~\citep{wei2022chainofthought}.

\vspace{-0.1cm}
\subsection{Performance of Global Agents}
\vspace{-0.1cm}
Table~\ref{tab:constraint_metrics} reports detection performance for seven global constraint types using the open-source LLM Phi4. We evaluate on all 567 instances from the \textsc{Text2Zinc} dataset~\citep{singirikonda2025text2zinc}. For each constraint type, the detection rate is computed over the subset of instances whose ground-truth model contains that constraint. As shown in the table, the average detection rates is around 70\% to 80\%. To estimate the false detection rate, we use 110 verified \textsc{Text2Zinc} instances and exclude those whose ground-truth model includes the target constraint. 

As shown Table~\ref{tab:constraint_metrics}, overall, our agents achieve strong detection rates, and false detection rates are generally low for rarer constraints; the main exception is \texttt{count} (28.3\%), suggesting that distinguishing counting patterns from other numerical constraints remains a challenge. These results indicate that the individual agents are reasonably effective, with room for improvement through stronger base models or prompt optimization.

\begin{table}[t] 
\centering 
\begin{tabular}{llccc} 
\toprule 
\textbf{Model} & \textbf{Strategy} & \textbf{Execution Rate (count)} & \textbf{Solve Rate (count)} & \textbf{Avg Score} \\ 
\midrule 
o3-mini & \textsc{Gala} & \textbf{57.27\% \; (63)} & \textbf{32.73\% \; (36)} & \textbf{45.00\%} \\
o3-mini & CoT & 52.73\% \; (58) & 30.91\% \; (34) & 41.82\% \\
\midrule 
gpt-4o-mini & \textsc{Gala} & \textbf{33.64\% \; (37)} & \textbf{17.27\% \; (19)} & \textbf{25.45\%}  \\ 
gpt-4o-mini & CoT & 31.82\% \; (35) & 12.73\% \; (14) & 22.27\% \\
\midrule 
gpt-oss:20b & \textsc{Gala} & \textbf{17.27\% \; (19)} & 8.18\% \; ( 9) & 12.73\%  \\ 
gpt-oss:20b & CoT & 16.36\% \; (18) & \textbf{10.00\% \; (11)} & \textbf{13.18\%}  \\ 
gpt-oss:20b & baseline& 11.81\% \; (13) & 7.27\% \; ( 8) & 9.54\%  \\
\bottomrule 
\end{tabular} 
\vspace{0.2cm}
\caption{Execution rate, solve rate (numbers of executed/solved instances in parentheses), and average score across 110 \textsc{Text2Zinc} instances.} 
\label{tab:current_performance_compact} 
\vspace{-0.6cm}
\end{table}

\vspace{-0.1cm}
\subsection{Performance of \textsc{Gala} on Text-to-Model Translation}
\vspace{-0.1cm}
In Table~\ref{tab:current_performance_compact}, we compare \textsc{Gala} with direct prompting (baseline) and chain-of-thought (CoT) across three LLM configurations: OpenAI o3-mini~\citep{openai2025o3}, gpt-4o-mini~\citep{openai2023gpt4}, and the open-source gpt-oss 20B model~\citep{openai2025gptoss}. We execute each model and strategy on 110 verified instances from the \textsc{Text2Zinc} datasets~\citep{singirikonda2025text2zinc}.

Agents consistently outperform CoT on the stronger models (o3-mini, GPT-4o-mini) across execution, solve, and mean score, and remain competitive on the 20B open model. While CoT gives a clear lift over the non-agent baseline on 20B, our first-pass \emph{global agents}, with no model tuning, prompt optimization, or hyperparameter search, match or slightly exceed CoT on the stronger models, and substantially improve execution success overall. This supports the claim that the gains come from the decomposition/agentic assembly itself rather than prompt or parameter scaling.

\vspace{-0.1cm}
\section{Future Work}


\vspace{-0.1cm}
\textbf{Optimize Global Agents:} Replace hand-crafted prompts with systematic optimization (automatic search, curated few-shot exemplars, adversarial negatives) and consider fine-tuning per global constraint to boost precision and recall. Pair these with compile-time snippet validation so each agent not only detects constraints correctly but also emits syntactically valid \textsc{MiniZinc}.

\textbf{Unblock the Assembler:} Add a supervisor to extract variables and objectives before delegation, or a post-hoc linker to unify names and deduplicate constraints. Build a systematic error taxonomy to map where agents and the assembler succeed or fail, driving targeted fixes and feedback loops that patch prompts or trigger regenerations, improving local correctness and global assembly coherence.

\textbf{Scale Evaluation:} Go beyond small models (e.g., Phi-4) by running stronger LLMs (e.g., GPT-4) and sweeping both open- and closed-weight models across sizes. Benchmark on datasets rich in global-constraint instances, currently 70\% of \textsc{Text2Zinc} lacks them, to better showcase the approach and reveal architectural headroom.

\vspace{-0.1cm}
\section{Conclusion}
\vspace{-0.1cm}
We introduce \textbf{\textsc{Gala}}: Global LLM Agents for Text-to-Model translation of satisfaction and optimization problems. We leverage specialized LLM agents to global constraints and an assembler to integrate them. Even in a first pass, without tuning, prompt engineering, or hyperparameter search, \textsc{Gala} outperforms a carefully curated CoT on stronger models and improve executability on the open 20B model, indicating that decomposition and agentic assembly are the key drivers of gains. The approach is modular and immediately extensible: finer handling of subtle constraints, compile-aware validation of snippets, and more robust staged assembly (conflict resolution, retry, and self-rectification) are natural next steps. With systematic error analysis and targeted feedback loops, we expect to turn this prototype into a robust and strong default for modeling co-pilots.

\section*{Acknowledgment}
The National Science Foundation (NSF) partially supported the research under grant \#2112533: "NSF Artificial Intelligence Research Institute for Advances in Optimization (AI4OPT)" and grant \#2346058: "NRT-AI: Integrating Artificial Intelligence and Operations Research Technologies".

\bibliographystyle{plainnat}
\bibliography{main}

\end{document}